\documentclass[journal]{IEEEtran}
\IEEEoverridecommandlockouts

\usepackage{cite}
\usepackage{amsmath,amssymb,amsfonts}
\usepackage{algorithmic}
\usepackage{graphicx}
\usepackage{textcomp}
\usepackage{xcolor}
\usepackage{booktabs}
\usepackage{multirow}
\usepackage{array}
\usepackage[colorlinks,linkcolor=red,anchorcolor=red,citecolor=blue]{hyperref}
\usepackage[lined,boxed,ruled]{algorithm2e}

\graphicspath{{./figs/}}

\def\BibTeX{{\rm B\kern-.05em{\sc i\kern-.025em b}\kern-.08em
    T\kern-.1667em\lower.7ex\hbox{E}\kern-.125emX}}

\begin{document}

\title{MMAO-Cls: Metabolic Multi-Agent Optimization for Joint Feature Selection and Classifier Tuning}

\author{Jinliang Xu$^{*}$ and Liping Ma%
\thanks{Jinliang Xu received the bachelor's and master's degrees from Beijing University of Posts and Telecommunications, Beijing, China, where he also received the Ph.D. degree in computer science in 2020. He is with the China Academy of Information and Communications Technology, Beijing, China. His current research interests include heuristic optimization, agent-based applications, and blockchain systems. His e-mail: jlxufly@gmail.com and xujinliang@caict.ac.cn.}%
\thanks{Liping Ma is with the Department of Disease Control and Prevention, The Seventh Medical Center of Chinese PLA General Hospital, Beijing, China; e-mail: lipingmaqzx@163.com.}}

\maketitle

\begin{abstract}
This paper studies whether the Metabolic Multi-Agent Optimizer (MMAO) can act as a credible outer-loop optimizer for classification model selection. We propose MMAO-Cls, a mixed-space realization in which each agent jointly encodes a binary feature mask and classifier hyperparameters, while private energy, communal budget, role drift, and lifecycle turnover are mapped to the accuracy-complexity tradeoff of wrapper learning. The implementation is strengthened by deriving feature-budget adaptation from feature-information priors and by regularizing validation reward with both subset compactness and train-validation overfitting gap. We evaluate MMAO-Cls on seven standard tabular benchmarks with three seeds each and compare it against RandomSearch, GA-lite, PSO-lite, and an endogenous no-sharing ablation. On the aggregate validation objective, MMAO-Cls ranks second ($0.9433$) behind GA-lite ($0.9446$). On held-out test performance, it reaches mean score $0.8882$, improving over RandomSearch ($0.8808$) and GA-lite ($0.8857$), remaining close to PSO-lite ($0.8874$) and the no-sharing ablation ($0.8900$), while using the most compact mean held-out feature subset among all compared methods (feature ratio $0.4881$). Pairwise tests show that these margins are not yet statistically significant. The results support MMAO's applicability to classification and compact mixed-space search, while the contribution of communal sharing to held-out performance remains unresolved.
\end{abstract}


\section{Introduction}
\IEEEPARstart{C}{lassification} pipelines often rely on design decisions that are not learned directly by the base classifier. Feature subset selection, hyperparameter tuning, and complexity control usually remain outer-loop optimization problems. These problems are mixed by nature: a candidate pipeline may include binary feature-inclusion decisions, continuous or ordinal hyperparameters, and a nontrivial tradeoff between predictive quality and subset compactness \cite{kohavi1997wrappers,guyon2003introduction,xue2016survey}. This places the task close to mixed-variable black-box optimization and algorithm configuration, where recent studies emphasize both landscape heterogeneity and the practical difficulty of fair low-budget comparison \cite{prager2024exploratory,raponi2023optimizing}.

This makes classification a useful testbed for the Metabolic Multi-Agent Optimizer (MMAO). MMAO is not a classifier; it is a search-control framework built around a private-public metabolic economy. Agents earn or lose private energy, donate part of successful gains to a communal pool, drift continuously between exploratory and exploitative roles, and are replaced when their search behavior stops paying for itself. The framework hypothesis is that heterogeneous search should emerge from this closed loop rather than from a stack of externally attached schedules.

The classification setting is an informative stress test for that hypothesis. Compared with standard continuous benchmarks, model selection for classification is mixed-space, evaluation-expensive, and vulnerable to overfitting. If MMAO is genuinely reusable across domains, it should coordinate feature-mask search and hyperparameter search under one explanatory controller while still producing sensible held-out performance.

This paper develops that idea through MMAO-Cls. Each agent carries a mixed representation whose discrete part determines a feature subset and whose continuous part determines classifier hyperparameters. Reward is based on validation utility, subset compactness, and a mild overfitting penalty. The resulting method is best understood as an outer optimization framework for classification pipelines rather than as a new predictive model. MMAO-Cls uses MMAO to optimize \emph{how a classifier is configured}, not to replace the classifier itself.

The resulting contribution is therefore methodological as much as empirical. The paper asks whether one closed-loop controller can coordinate heterogeneous outer-loop decisions in a way that remains interpretable, compact, and reusable across classifier families.

The paper makes four contributions.
\begin{itemize}
    \item It defines a classification-specific derivation of MMAO in which feature masks, hyperparameter vectors, compactness pressure, and overfitting control are all coupled to one metabolic loop.
    \item It formulates a mixed objective that supports wrapper feature selection and classifier tuning within one run while exposing the underlying accuracy-complexity tradeoff explicitly.
    \item It implements a seven-dataset benchmark with shared encoding, shared objective, multiple lightweight baselines, and an endogenous no-sharing ablation.
    \item It positions MMAO conservatively in classification: as a credible outer-loop optimizer for mixed model configuration, not as a claim of universal superiority.
\end{itemize}

\section{Related Work}
Metaheuristic classification studies usually appear in two outer-loop forms. The first is wrapper feature selection, where a search process proposes subsets and a downstream classifier evaluates them \cite{kohavi1997wrappers,guyon2003introduction,xue2016survey}. The second is hyperparameter optimization, where the metaheuristic explores model configurations such as support-vector-machine kernels, neighborhood size for $k$-nearest neighbors, or regularization for linear models \cite{bergstra2012random,feurer2019hyperparameter}. In both cases, the classifier itself remains standard while the heuristic acts as a configuration layer around it. More recent mixed-variable and low-budget black-box optimization studies reinforce that this outer-loop view is a legitimate optimization problem in its own right rather than a mere implementation detail \cite{prager2024exploratory,raponi2023optimizing,nesterov2017random}.

We examine whether MMAO can provide a mixed-space optimizer with metabolically interpretable adaptation. This connects the paper to broader work on self-adaptation, parameter control, and adaptive operator or strategy selection in evolutionary computation \cite{Hansen2001,eiben1999parameter,brest2006self,karafotias2015parameter,li2014adaptivebandit,omeradzic2024self,jiang2023knowledge,cho2025configx}. It also connects to resource-allocation research, where search effort is redistributed across populations, subproblems, or tasks according to recent utility signals \cite{li2022distributed,liu2022cooperative,dong2025effective}. Finally, it connects to current community pressure for clearer benchmarking, mechanism analysis, and behavior-level comparison in metaphor-based optimization \cite{vermetten2024large,stripinis2024benchmarking,cenikj2025comparing}.

\section{From MMAO to MMAO-Cls}
\subsection{Design Principle}
The same framework constraint used in earlier MMAO framework studies is retained here \cite{xu2026mmao,xu2026minimalmmao}:
\begin{quote}
all adaptive behavior in MMAO-Cls should be explainable as a consequence of the metabolic resource loop rather than as an external controller bolted onto a classification wrapper.
\end{quote}

This matters because classification optimization can involve many separate control rules. One may add separate schedules for feature mutation, parameter tuning, survivor selection, sparsity repair, and population resizing. MMAO-Cls instead treats these as consequences of one economy of search effort.

\subsection{Mixed Agent State}
For a dataset with $d$ original features, each agent is represented by
\begin{equation}
    A_i(t)=\big(z_i(t), \theta_i(t), E_i(t), \phi_i(t), m_i(t)\big),
\end{equation}
where $z_i(t)\in\{0,1\}^d$ is a binary feature mask, $\theta_i(t)$ is a continuous or ordinal hyperparameter vector, $E_i(t)$ is private energy, $\phi_i(t)\in[0,1]$ is role state, and $m_i(t)$ stores the best mixed configuration previously discovered by the agent.

The population additionally maintains a communal budget $B_t$ and a recent gain scale $s_t$. As in the core MMAO formulation \cite{xu2026mmao}, progress is normalized before it is rewarded, so that different datasets and classifiers do not create incomparable raw score magnitudes. This normalization is especially important in mixed-variable search, where heterogeneous subspaces can create very different local sensitivities and reward scales \cite{prager2024exploratory}.

\subsection{Feature-Prior and Budget Mapping}
The classification version uses feature-information priors derived from mutual information on the training split. Let $\pi_j \ge 0$ denote the normalized prior score of feature $j$. The priors are not used as a hard filter. Instead, they shape the endogenous target subset size through two cumulative-information thresholds:
\begin{equation}
    k_{\min}=\min \left\{k:\sum_{j \in \text{top-}k}\pi_j \ge 0.68 \sum_{j=1}^{d}\pi_j \right\},
\end{equation}
\begin{equation}
    k_{\mathrm{tar}}=\min \left\{k:\sum_{j \in \text{top-}k}\pi_j \ge 0.90 \sum_{j=1}^{d}\pi_j \right\},
\end{equation}
with a small lower bound for very low-dimensional tasks. The corresponding target ratio is $\rho^\star = k_{\mathrm{tar}}/d$. This turns sparsity control into a resource target inferred from the data rather than a fixed externally chosen subset size.

\subsection{Objective Mapping}
Let $Q_{\mathrm{val}}(z,\theta)$ denote validation utility under a fixed train-validation protocol, and let
\begin{equation}
    r(z)=\frac{\|z\|_0}{d}
\end{equation}
be the selected-feature ratio. The implemented MMAO-Cls objective is
\begin{equation}
    J(z,\theta)=Q_{\mathrm{val}}(z,\theta)-\lambda_f \, g(r(z),\rho^\star)-\lambda_o \, \Delta_{\mathrm{overfit}},
\end{equation}
where $\lambda_f=0.16$, $\lambda_o=0.18$,
\begin{equation}
    \Delta_{\mathrm{overfit}}=\max \big(0, Q_{\mathrm{train}}-Q_{\mathrm{val}}\big),
\end{equation}
and the compactness term is activated only when the current subset exceeds the target ratio:
\begin{equation}
    g(r,\rho^\star)=0.85 \max(0,r-\rho^\star)+0.12 \max(0,r-\rho^\star)^2.
\end{equation}
This mapping is intentionally simple. One run simultaneously supports wrapper feature selection and hyperparameter tuning, but the reported validation metric and feature ratio still expose the underlying bi-criteria structure.

\subsection{Metabolic Interpretation}
The classification reinterpretation of the MMAO loop is summarized as follows.
\begin{itemize}
    \item \textbf{Private energy:} tracks whether a mixed configuration continues to yield useful predictive-complexity gains.
    \item \textbf{Communal budget:} stores socialized improvement and finances new search around high-value encodings.
    \item \textbf{Role drift:} shifts agents along a continuum between broader mask perturbation and narrower hyperparameter refinement.
    \item \textbf{Lifecycle turnover:} removes low-yield agents and respawns search where the current economy can afford it.
    \item \textbf{Memory and consensus:} preserve successful mixed encodings and expose their shared mask structure for later mutation.
\end{itemize}

\subsection{Operational Details}
The current implementation instantiates this mapping with five practical mechanisms.
\begin{itemize}
    \item Feature logits are initialized and perturbed with a prior pull toward informative features.
    \item Elite consensus over historical masks provides a soft social signal instead of a hard template.
    \item Candidate evaluation includes mask repair so that extremely under-complete subsets are corrected before classifier fitting.
    \item A local elite-refinement step may drop weak selected features or add high-prior unselected features if this does not damage the metabolic objective.
    \item Role updates depend on stability and current subset ratio, so exploration pressure and refinement pressure remain coupled to search outcome.
\end{itemize}

\begin{figure}[t]
\centering
\includegraphics[width=\columnwidth]{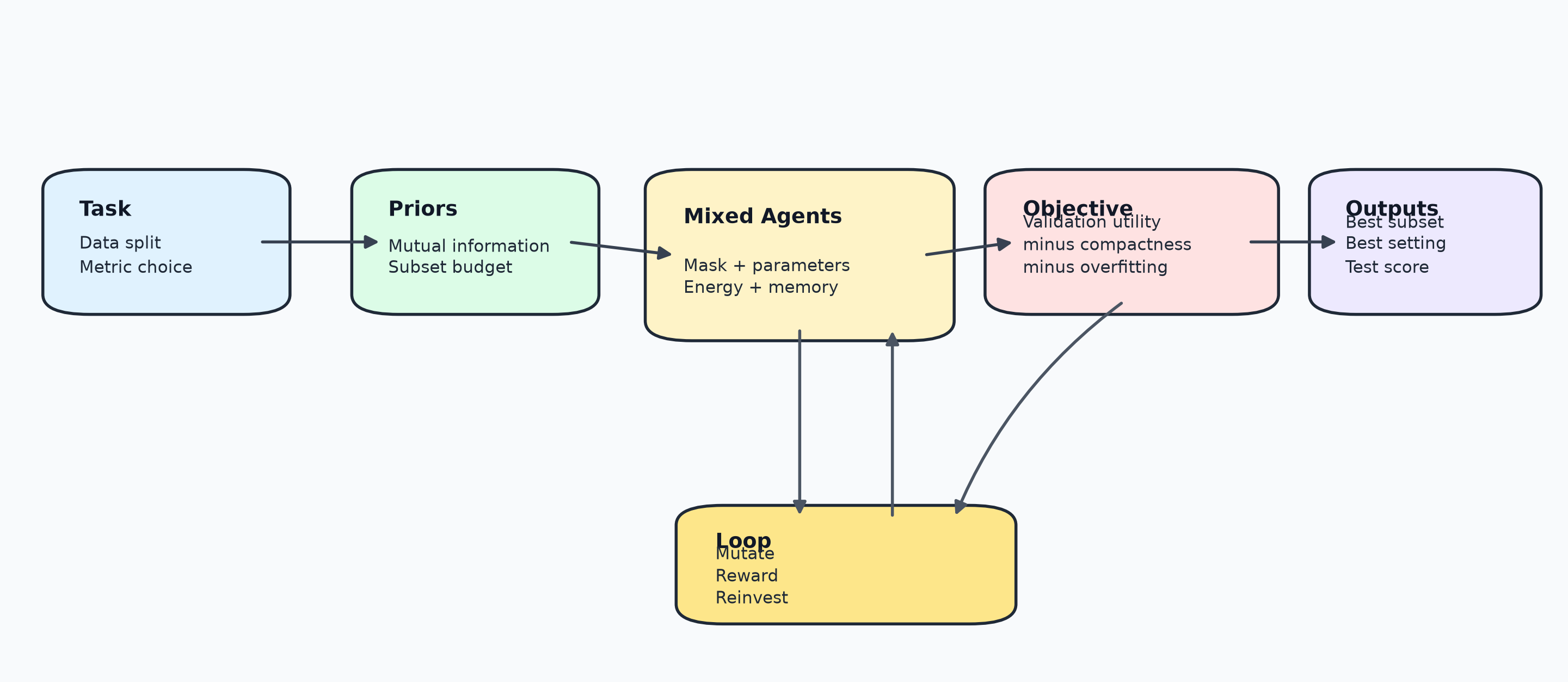}
\caption{Workflow of MMAO-Cls. Dataset structure and feature priors feed a mixed agent population, whose search behavior is updated through one closed metabolic loop linking evaluation, resource redistribution, and mixed-space refinement.}
\label{fig:workflow}
\end{figure}

Figure~\ref{fig:workflow} summarizes this mapping at a systems level. Classifier choice, feature priors, mixed-state mutation, and compactness-aware evaluation are coupled through one resource loop in which validation gain, subset pressure, and communal reinvestment shape later search behavior.

\subsection{Algorithmic Skeleton}
\begin{algorithm}[t]
\caption{High-level MMAO-Cls}
\label{alg:mmaocls}
\DontPrintSemicolon
Initialize mixed agents with feature logits, hyperparameters, energies, and role states\;
split the dataset into train, validation, and held-out test subsets\;
compute feature-information priors and derive target subset budget\;
evaluate each agent on the train-validation objective with compactness and overfitting penalties\;
\For{$t=1,\dots,T$}{
    update the recent gain scale and communal budget\;
    \ForEach{agent}{
        generate a mixed candidate by perturbing feature logits and hyperparameters according to role, priors, consensus, and anchor memory\;
        repair extremely under-complete masks and perform a local add/drop elite refinement\;
        evaluate predictive utility and feature ratio on the validation split\;
        reward normalized improvement, update private energy, and donate the communal share\;
        shift the role state between broader subset exploration and finer hyperparameter refinement\;
    }
    prune weak agents and respawn if the population falls below the minimum\;
    if surplus communal budget exists, spawn new candidates around high-value mixed encodings\;
    record the best objective, predictive metric, and selected-feature ratio\;
}
\Return the best mixed configuration together with its held-out test evaluation\;
\end{algorithm}

\section{Experimental Design}
\subsection{Datasets and Classifiers}
The current benchmark contains seven standard tabular datasets (see GitHub repository \texttt{mmao}\footnote{\url{https://github.com/wolfbrother/mmao}} or PyPI package \texttt{mmao-opt}\footnote{\url{https://pypi.org/project/mmao-opt/}}). Four come from the classical \texttt{scikit-learn} collection: \texttt{breast\_cancer}, \texttt{wine}, \texttt{digits}, and \texttt{iris}. Three additional datasets are obtained through the standard OpenML/UCI access path used by \texttt{scikit-learn}: \texttt{sonar}, \texttt{ionosphere}, and \texttt{vehicle}. This mix gives a moderate spread in dimensionality, class structure, and difficulty while remaining computationally manageable for repeated train-validation-test evaluation.

Three classifier families are used:
\begin{itemize}
    \item RBF support vector machines for \texttt{breast\_cancer}, \texttt{wine}, and \texttt{sonar};
    \item $k$-nearest neighbors for \texttt{digits} and \texttt{iris};
    \item logistic regression for \texttt{ionosphere} and \texttt{vehicle}.
\end{itemize}

This choice is deliberate. All three families are classical, well understood, and sensitive to feature selection and hyperparameter configuration, which makes them suitable for testing a classification-oriented outer optimizer without introducing heavy model-training overhead. It also keeps the benchmark closer to the limited-budget model-selection setting emphasized in recent black-box optimization studies \cite{raponi2023optimizing}.

\subsection{Protocol and Metrics}
For each seed in $\{3,7,11\}$, data are split into train, validation, and held-out test partitions in a stratified $56.25\%/18.75\%/25\%$ ratio. Balanced accuracy is used for \texttt{breast\_cancer}, \texttt{wine}, \texttt{sonar}, and \texttt{ionosphere}; macro-F1 is used for \texttt{digits}, \texttt{iris}, and \texttt{vehicle}. All classifier evaluations are carried out in training-derived preprocessing pipelines, with feature-information priors computed only from the training split and standard scaling fit only within the corresponding training data. Optimization is performed on the train-validation objective $J(z,\theta)$, but we report four views:
\begin{itemize}
    \item objective value;
    \item raw validation metric;
    \item selected-feature ratio;
    \item held-out test metric.
\end{itemize}
This separation is important because a method may improve the scalar objective by improving predictive quality, by using fewer features, or by both.

\subsection{Baselines and Fairness}
The comparison set contains:
\begin{itemize}
    \item \texttt{RandomSearch}: mixed-space random sampling over the same encoding;
    \item \texttt{GA-lite}: a lightweight evolutionary baseline with crossover and mutation;
    \item \texttt{PSO-lite}: a lightweight swarm baseline with personal and global best guidance;
    \item \texttt{NoResourceSharing}: an endogenous ablation that removes communal sharing while retaining the MMAO-style local dynamics.
\end{itemize}
All methods share the same candidate encoding, feature priors, classifier family, seed set, and optimization objective. The main fairness constraint is therefore representational rather than fully evaluation-matched. Every method uses $24$ outer iterations, and non-MMAO baselines use $10$ candidates per iteration. MMAO-Cls starts from $10$ agents and can contract or grow within the range $[6,20]$ through its endogenous population dynamics. This kind of endogenous population control is consistent with prior work showing that search quality can depend strongly on how population size is adapted rather than fixed \cite{Tanabe2014,liu2025less,doerr2025speeding,antipov2024already}. Accordingly, the present study is best read as a mechanism-validation benchmark with aligned outer-loop schedules, not as a final function-evaluation-normalized competition, a distinction also stressed in recent benchmarking literature \cite{stripinis2024benchmarking,raponi2023optimizing}.

\section{Results}
\subsection{Aggregate Reading}
Table~\ref{tab:aggregate} summarizes the seven-dataset, three-seed benchmark. On the optimization objective, MMAO-Cls reaches mean score $0.9433$, second only to GA-lite at $0.9446$. On held-out test performance, MMAO-Cls reaches $0.8882$, ahead of RandomSearch ($0.8808$) and GA-lite ($0.8857$), very close to PSO-lite ($0.8874$), and slightly below the no-sharing ablation ($0.8900$). The strongest aggregate advantage of MMAO-Cls is compactness: it uses the lowest mean feature ratio both on validation ($0.4923$) and on held-out test ($0.4881$).

MMAO-Cls is competitive on the joint objective and reasonably strong on held-out accuracy, but the margins are small. The classification evidence is therefore stronger for compact mixed-space search than for decisive superiority.

\begin{table*}[t]
\caption{Aggregate results across seven datasets and three seeds. Higher objective and metric values are better; lower feature ratios are better.}
\label{tab:aggregate}
\centering
\small
\resizebox{\textwidth}{!}{%
\begin{tabular}{lcccccc}
\toprule
Method & Mean objective & Mean validation metric & Mean feature ratio & Mean test metric & Avg. objective rank & Avg. test rank \\
\midrule
GA-lite & 0.9446 & 0.9481 & 0.5413 & 0.8857 & 2.214 & 3.143 \\
MMAO-Cls & 0.9433 & 0.9466 & 0.4923 & 0.8882 & 2.643 & 3.143 \\
NoResourceSharing & 0.9413 & 0.9440 & 0.5039 & 0.8900 & 2.929 & 2.571 \\
PSO-lite & 0.9414 & 0.9444 & 0.5471 & 0.8874 & 3.357 & 2.714 \\
RandomSearch & 0.9371 & 0.9402 & 0.5309 & 0.8808 & 3.857 & 3.429 \\
\bottomrule
\end{tabular}
}
\end{table*}

\subsection{Accuracy-Compactness Reading}
\begin{figure}[t]
\centering
\includegraphics[width=\columnwidth]{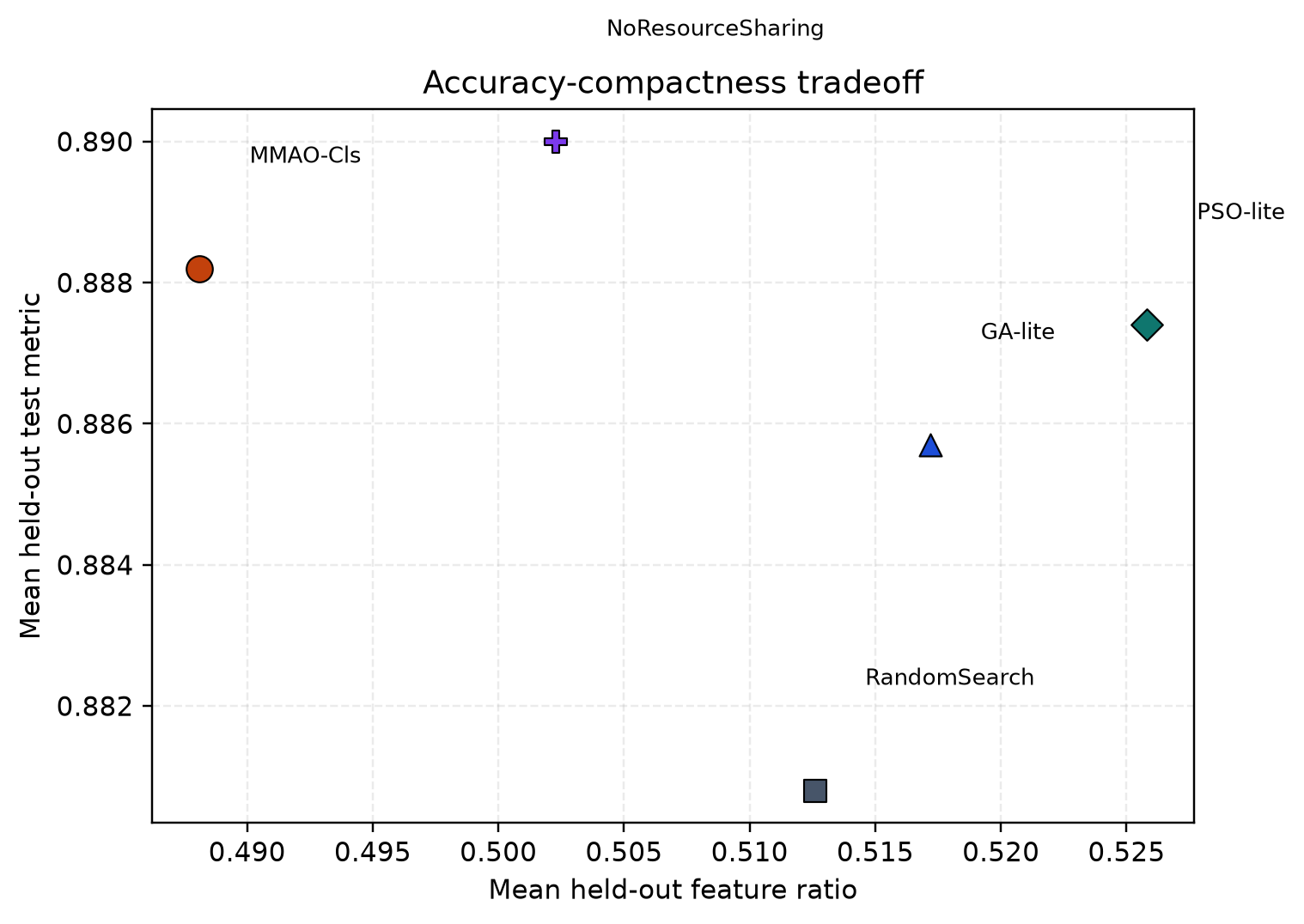}
\caption{Aggregate held-out accuracy-compactness tradeoff. MMAO-Cls remains close to the best test metric group while using the smallest mean held-out feature ratio.}
\label{fig:tradeoff}
\end{figure}

The most stable classifier-side advantage of MMAO-Cls is clearer in the accuracy-compactness plane than in raw test score alone. Figure~\ref{fig:tradeoff} plots the aggregate held-out test metric against the aggregate held-out feature ratio. MMAO-Cls occupies the most compact position among all methods while remaining in the top group of test performance. Table~\ref{tab:compactness} makes the same point numerically: the held-out efficiency ratio $\text{test\_metric}/\text{test\_feature\_ratio}$ is highest for MMAO-Cls ($1.8197$), above NoResourceSharing ($1.7718$), RandomSearch ($1.7183$), GA-lite ($1.7125$), and PSO-lite ($1.6877$). These results indicate that MMAO-Cls tends to find competitive predictive configurations with smaller feature subsets.

\begin{table}[t]
\caption{Held-out compactness summary. Larger efficiency means more predictive performance per unit feature ratio.}
\label{tab:compactness}
\centering
\small
\resizebox{\columnwidth}{!}{%
\begin{tabular}{lccc}
\toprule
Method & Test metric & Test feature ratio & Efficiency \\
\midrule
MMAO-Cls & 0.8882 & 0.4881 & 1.8197 \\
NoResourceSharing & 0.8900 & 0.5023 & 1.7718 \\
RandomSearch & 0.8808 & 0.5126 & 1.7183 \\
GA-lite & 0.8857 & 0.5172 & 1.7125 \\
PSO-lite & 0.8874 & 0.5258 & 1.6877 \\
\bottomrule
\end{tabular}
}
\end{table}

\subsection{Dataset-Level Structure}
Table~\ref{tab:dataset-summary} shows how this picture decomposes by dataset. MMAO-Cls produces the best objective on \texttt{breast\_cancer} and ties for best objective on \texttt{wine}; it also delivers strong test performance on \texttt{digits}, \texttt{sonar}, and \texttt{wine} with comparatively compact subsets. The weak points are \texttt{ionosphere} and \texttt{vehicle}, where the present metabolic controller does not convert its compactness bias into better held-out performance. The easiest tasks, especially \texttt{iris}, remain too small to differentiate the methods sharply.

\begin{table*}[t]
\caption{Dataset-level summary. ``Best objective'' and ``best test'' report the strongest mean among all methods on that dataset.}
\label{tab:dataset-summary}
\centering
\small
\resizebox{\textwidth}{!}{%
\begin{tabular}{lccccc}
\toprule
Dataset & MMAO objective & Best objective & MMAO test metric & Best test metric & MMAO feature ratio \\
\midrule
Breast Cancer & 0.9909 & 0.9909 (MMAO-Cls) & 0.9445 & 0.9695 (RandomSearch) & 0.4333 \\
Digits & 0.9891 & 0.9914 (GA-lite) & 0.9792 & 0.9829 (NoResourceSharing) & 0.4792 \\
Ionosphere & 0.9001 & 0.9102 (GA-lite) & 0.8274 & 0.8475 (GA-lite) & 0.5784 \\
Iris & 0.9876 & 1.0000 (RandomSearch) & 0.9385 & 0.9472 (PSO-lite) & 0.3333 \\
Sonar & 0.9397 & 0.9523 (PSO-lite) & 0.7927 & 0.7986 (NoResourceSharing) & 0.3056 \\
Vehicle & 0.7960 & 0.8102 (GA-lite) & 0.7565 & 0.7785 (PSO-lite) & 0.7778 \\
Wine & 1.0000 & 1.0000 (tie) & 0.9790 & 0.9790 (tie) & 0.5385 \\
\bottomrule
\end{tabular}
}
\end{table*}

\subsection{Cross-Classifier Reuse}
An important part of the MMAO claim is that one controller can be reused across different classifier families without being redesigned for each one. Table~\ref{tab:classifier-family} summarizes this cross-family reading. On the \texttt{svm\_rbf} tasks, MMAO-Cls attains the best mean held-out test metric ($0.9054$) while remaining the second most compact method ($0.4258$). On the \texttt{knn} tasks, MMAO-Cls again stays in the top performance group ($0.9588$) and becomes the most compact method by a clear margin ($0.4062$). The \texttt{logreg} tasks show weaker performance, although MMAO-Cls remains competitive. This cross-family stability strengthens the interpretation of MMAO-Cls as a reusable outer-loop controller rather than as a one-classifier wrapper.

\begin{table*}[t]
\caption{Classifier-family summary. Ranks are computed within each classifier family; smaller compactness rank means fewer features.}
\label{tab:classifier-family}
\centering
\small
\resizebox{\textwidth}{!}{%
\begin{tabular}{lccccccc}
\toprule
Classifier & Method & Objective & Test metric & Feature ratio & Objective rank & Test rank & Compactness rank \\
\midrule
knn & MMAO-Cls & 0.9883 & 0.9588 & 0.4062 & 2 & 2 & 1 \\
knn & NoResourceSharing & 0.9876 & 0.9607 & 0.4271 & 3 & 1 & 2 \\
knn & RandomSearch & 0.9889 & 0.9566 & 0.4896 & 1 & 5 & 3 \\
logreg & MMAO-Cls & 0.8480 & 0.7920 & 0.6781 & 2 & 3 & 4 \\
logreg & NoResourceSharing & 0.8444 & 0.7964 & 0.7119 & 4 & 2 & 5 \\
logreg & GA-lite & 0.8602 & 0.8125 & 0.6645 & 1 & 1 & 2 \\
svm\_rbf & MMAO-Cls & 0.9768 & 0.9054 & 0.4258 & 2 & 1 & 2 \\
svm\_rbf & NoResourceSharing & 0.9751 & 0.9053 & 0.4165 & 3 & 2 & 1 \\
svm\_rbf & PSO-lite & 0.9777 & 0.9037 & 0.4855 & 1 & 3 & 5 \\
\bottomrule
\end{tabular}
}
\end{table*}

\subsection{Pairwise and Statistical Reading}
To reduce over-interpretation of small mean gaps, Table~\ref{tab:pairwise} reports paired test-set comparisons over the $21$ dataset-seed pairs. MMAO-Cls records positive mean differences against RandomSearch, GA-lite, and PSO-lite, and a small negative mean difference against NoResourceSharing. However, all paired Wilcoxon tests remain non-significant ($p \ge 0.4980$). This confirms that the present benchmark supports a competitiveness claim more strongly than a dominance claim.

\begin{table}[t]
\caption{Paired held-out test comparison for MMAO-Cls over $21$ dataset-seed pairs.}
\label{tab:pairwise}
\centering
\small
\resizebox{\columnwidth}{!}{%
\begin{tabular}{lccc}
\toprule
Baseline & Win-Loss-Tie & Mean $\Delta$ test & Wilcoxon $p$ \\
\midrule
RandomSearch & 12-7-2 & 0.0074 & 0.5153 \\
GA-lite & 13-7-1 & 0.0026 & 0.4980 \\
PSO-lite & 9-11-1 & 0.0009 & 0.9854 \\
NoResourceSharing & 5-7-9 & -0.0018 & 0.7334 \\
\bottomrule
\end{tabular}
}
\end{table}

\subsection{Mechanism Diagnostics}
The mechanism-level summaries remain informative even when the final test margins are small. MMAO-Cls ends with mean population size $20$, mean communal budget $11.0124$, and mean final feature ratio $0.4921$. By contrast, the no-sharing ablation ends with mean budget $1.0000$ and a slightly larger mean feature ratio $0.5011$. GA-lite, PSO-lite, and RandomSearch all use larger average subsets. This suggests that the metabolic loop maintains shared resources and encourages more compact subsets.

\begin{figure*}[t]
\centering
\includegraphics[width=\textwidth]{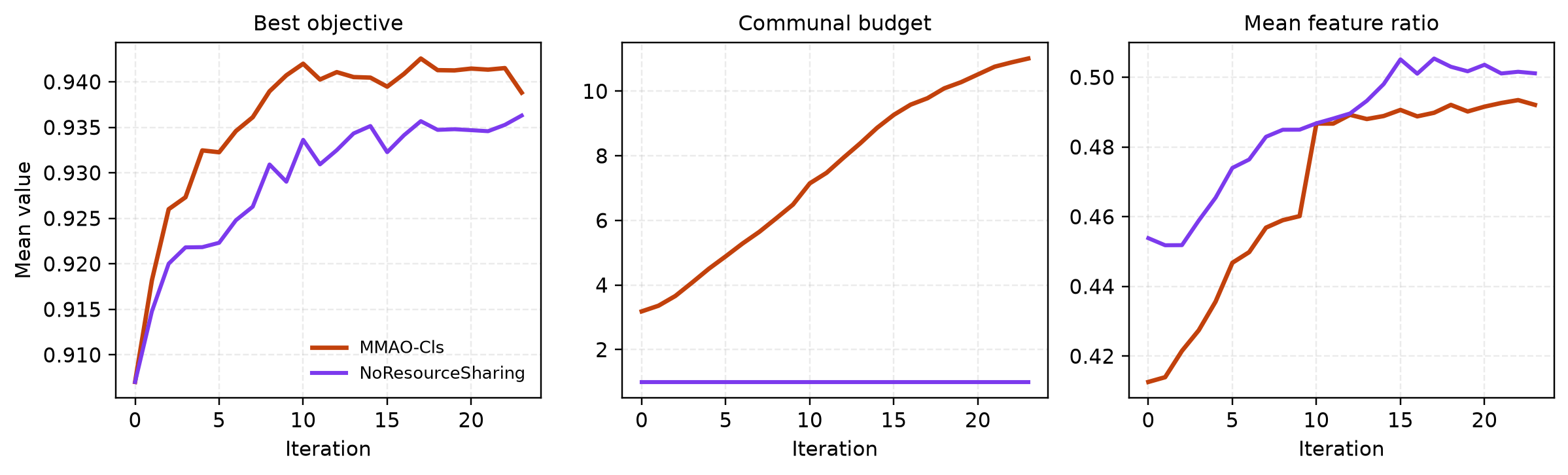}
\caption{Mean trajectory comparison between MMAO-Cls and NoResourceSharing over all runs. The full method accumulates communal budget, improves the objective slightly faster, and reaches a slightly more compact mean feature ratio.}
\label{fig:mechanism-compare}
\end{figure*}

Figure~\ref{fig:mechanism-compare} makes this contrast more concrete by comparing the mean histories of MMAO-Cls and NoResourceSharing. Three differences stand out. First, the best objective of MMAO-Cls rises faster and stabilizes at a slightly better level. Second, the communal budget of MMAO-Cls grows steadily, whereas the no-sharing ablation stays flat by construction. Third, the mean feature ratio of MMAO-Cls is pushed downward earlier in the run and remains slightly more compact afterward. The shared budget thus produces a different search trajectory. What the current evidence still does \emph{not} prove is that this trajectory difference translates reliably into superior held-out accuracy on every benchmark family.

\begin{table}[t]
\caption{Mechanism summary averaged over all runs.}
\label{tab:mechanism}
\centering
\small
\resizebox{\columnwidth}{!}{%
\begin{tabular}{lccc}
\toprule
Method & Final population & Final communal budget & Final mean feature ratio \\
\midrule
MMAO-Cls & 20.0000 & 11.0124 & 0.4921 \\
NoResourceSharing & 10.0000 & 1.0000 & 0.5011 \\
GA-lite & 10.0000 & 0.0000 & 0.5413 \\
PSO-lite & 10.0000 & 0.0000 & 0.5471 \\
RandomSearch & 10.0000 & 0.0000 & 0.5309 \\
\bottomrule
\end{tabular}
}
\end{table}

\subsection{What the Current Evidence Supports}
Taken together, the current results support four restrained conclusions.
\begin{itemize}
    \item MMAO can be instantiated credibly in classification as an outer optimizer for joint feature selection and hyperparameter tuning.
    \item The current MMAO-Cls implementation is strongest when the target problem rewards compact mixed configurations without requiring very aggressive classifier-specific adaptation.
    \item The metabolic controller appears to encourage compact subsets and persistent search capital, but the current evidence does not show that communal sharing is already a decisive source of held-out-performance gains.
    \item The benchmark supports classification applicability, while broader comparisons with specialist wrappers and AutoML-style optimizers remain necessary.
\end{itemize}

\section{Discussion and Positioning}
MMAO-Cls should be interpreted carefully. It is not a proposal to replace standard classifiers, and it is not aimed at large-scale deep learning settings where model training is already dominated by gradient information and expensive architecture choices. Its natural domain is outer-loop configuration: mixed feature selection, hyperparameter tuning, complexity control, and structurally heterogeneous search. In that sense, it is closer to mixed-variable algorithm configuration than to end-to-end AutoML, and its most relevant comparators are resource-aware derivative-free optimizers and configurable outer-loop controllers \cite{prager2024exploratory,raponi2023optimizing,cho2025configx}.

The classification results provide further evidence of transfer across optimization domains. The present results are consistent with that reading: the algorithm is competitive, compact, and mechanistically interpretable, but its metabolic advantages are still partial rather than overwhelming. This restrained interpretation is important because recent benchmark analyses increasingly caution against reading small aggregate gaps as evidence of a new universally superior heuristic family \cite{vermetten2024large,stripinis2024benchmarking,cenikj2025comparing}.

\subsection{Why the Classification Mapping Is Meaningful}
Two properties of the present results matter more than the exact ranking of means.
\begin{itemize}
    \item MMAO-Cls remains competitive while using the smallest average feature subset among all compared methods.
    \item The same controller handles binary mask search and continuous or ordinal hyperparameter search without introducing a separate optimizer for each subspace.
\end{itemize}
This is the main sense in which the classification extension works. The method does not merely tune a classifier; it coordinates two structurally different search subproblems through one budget loop. In the present paper, that claim should be read as evidence of reusable mixed-space control, not as evidence that MMAO-Cls has already become a best-in-class classification optimizer.

Classification is also a particularly stringent transfer setting for MMAO. Unlike pure continuous search or pure combinatorial routing, the wrapper must navigate a mixed discrete-continuous configuration space, tolerate noisy validation feedback, and avoid spending its budget on feature inflation that does not survive held-out testing. Maintaining coherent control under these conditions tests the framework's ability to coordinate heterogeneous representations, evaluation cost, and overfitting risk.

\subsection{Where the Present MMAO-Cls Still Falls Short}
The present evidence is equally clear about what remains unresolved.
\begin{itemize}
    \item Communal sharing is not yet cleanly vindicated, because the no-sharing ablation slightly exceeds MMAO-Cls on aggregate held-out test score.
    \item Some datasets, especially \texttt{ionosphere} and \texttt{vehicle}, suggest that the current compactness bias can still become too conservative or misaligned with classifier behavior.
    \item The baseline set is broader than a demo-level study, but it is still lighter than a full AutoML or specialist-wrapper comparison.
    \item Evaluation budgets are aligned by outer-loop schedule rather than exactly matched by function evaluations, which matters because MMAO can grow its population endogenously.
 \end{itemize}

The mechanism reading should therefore stay selective. The clearest present benefit of the metabolic loop is compactness-oriented resource use: the controller tends to preserve search capital for smaller, reusable mixed configurations rather than only chasing marginal validation gains through larger masks. By contrast, communal sharing is not yet cleanly vindicated as a decisive held-out accuracy mechanism, because the no-sharing ablation remains extremely close. This is still a useful result, since it tells us which part of the framework currently carries the strongest practical value in classification.

\subsection{Limitations and Future Work}
Several limitations are especially important.
\begin{itemize}
    \item The benchmark is still moderate in scale: seven datasets, three seeds, and lightweight but reproducible baselines.
    \item The objective is scalarized; a genuine multi-objective MMAO-Cls could model accuracy and feature count as an explicit Pareto set.
    \item The communal-sharing benefit is not yet cleanly separated from the rest of the mixed-search design.
    \item The present study does not compare against stronger configuration tools such as Bayesian optimization, SMAC-style configurators, or learned configuration agents \cite{raponi2023optimizing,cho2025configx}.
    \item Exact function-evaluation matching remains to be done for stronger claims about search efficiency, especially when dynamic population sizing is part of the algorithmic identity \cite{Tanabe2014,doerr2025speeding}.
\end{itemize}

Natural next steps therefore include larger OpenML coverage, stronger outer-loop baselines, exact budget normalization, explicit Pareto analysis for accuracy versus feature count, and broader classifier families such as tree ensembles or lightweight neural models. A second direction is stronger mechanism diagnosis: mixed-variable landscape characterization, behavior-level search comparison, and more explicit study of when adaptive resource sharing helps more than classical self-adaptation alone \cite{prager2024exploratory,cenikj2025comparing,brest2006self,liu2025less}. A third is stronger reproducibility packaging, including frozen benchmark manifests and exact evaluation-count accounting for dynamic-population runs.

\section{Conclusion}
This paper proposes MMAO-Cls, a classification-oriented MMAO derivative for joint feature selection and classifier hyperparameter tuning. The central idea is to map mixed discrete-continuous model selection into the same metabolic resource loop that underlies MMAO elsewhere, so that predictive utility, subset compactness, overfitting pressure, role adaptation, and candidate turnover are governed by one endogenous controller. On the present seven-dataset benchmark, MMAO-Cls is competitive on the optimization objective, improves over RandomSearch and GA-lite on mean held-out test score, and uses the most compact mean held-out feature subset among all compared methods. The additional tradeoff and classifier-family analyses sharpen that conclusion: the most convincing current strength of MMAO-Cls is not absolute predictive dominance but compact mixed-space model selection under one reusable controller. At the same time, the margins are small and the no-sharing ablation remains extremely close. The results support MMAO's transfer to classification as a compact mixed-space outer optimization framework, with further work needed to establish the benefit of communal resource sharing.

More broadly, the paper shows that classification can serve as a demanding transfer domain for MMAO rather than merely as an application vignette. Mixed representation, evaluation noise, and compactness pressure make it a useful setting for testing whether the framework remains coherent outside standard benchmark landscapes.

Accordingly, the paper should be read as evidence of cross-domain transferability and mixed-space controller reuse, rather than as a claim that MMAO-Cls is already a finished alternative to specialist AutoML or wrapper-optimization systems.

\section*{Acknowledgement}
The related publications in Table~\ref{tab:mmao-research-series} provide complementary views of the MMAO research series.

\begin{table}[!ht]
\centering
\caption{Other publications in the MMAO research series.}
\label{tab:mmao-research-series}
\footnotesize
\setlength{\tabcolsep}{5pt}
\renewcommand{\arraystretch}{1.12}
\begin{tabular}{@{}p{\columnwidth}@{}}
\hline
\textbf{Related publication} \\
\hline
\emph{MMAO: A Metabolic Multi-Agent Optimizer with Endogenous Resource Allocation for Continuous and Discrete Optimization}~\cite{xu2026mmao}.\newline
\href{https://arxiv.org/abs/2606.28109}{arXiv:2606.28109}
\\[5pt]
\emph{Minimal MMAO: A Resource-Closed-Loop Framework for Adaptive Metaheuristic Search}~\cite{xu2026minimalmmao}.\newline
\href{https://arxiv.org/abs/2606.30450}{arXiv:2606.30450}
\\[5pt]
\emph{A Large-Scale Empirical Evaluation of MMAO Under Fair-Budget Continuous and Discrete Benchmarks}~\cite{xu2026largescale}.\newline
\href{https://arxiv.org/abs/2606.31584}{arXiv:2606.31584}
\\[5pt]
\emph{MMAO-Dyn: A Metabolic Multi-Agent Optimizer for Dynamic Optimization}~\cite{xu2026mmaodyn}.\newline
\href{https://arxiv.org/abs/2607.00846}{arXiv:2607.00846}
\\[5pt]
\emph{Mechanism and Stability Analysis of Metabolic Closed-Loop Metaheuristics}~\cite{xu2026mechanism}.\newline
\href{https://arxiv.org/abs/2607.01551}{arXiv:2607.01551}
\\[5pt]
\emph{MMAO: A Unified Resource-Regulated Controller for Cross-Domain Optimization}~\cite{xu2026unifiedmmao}.\newline
\href{https://doi.org/10.1109/ACCESS.2026.3752083}{IEEE Access, 2026}
\\[5pt]
\hline
\end{tabular}
\end{table}

\bibliographystyle{IEEEtran}
\bibliography{Ref}

\end{document}